\documentclass[11pt]{article}

\usepackage{acl2012}
\usepackage{times}
\usepackage{latexsym}
\usepackage{amsmath}
\usepackage{multirow}
\usepackage{url}
\usepackage{amssymb}            
\usepackage{enumitem}           
\usepackage{graphicx}           
\usepackage{pbox}               
\usepackage{tikz}               

\title{Distributional Semantics Beyond Words: \\
       Supervised Learning of Analogy and Paraphrase}

\author{Peter D. Turney \\
  National Research Council Canada \\
  Information and Communications Technologies \\
  Ottawa, Ontario, Canada, K1A 0R6 \\
  {\tt peter.turney@nrc-cnrc.gc.ca} \\}

\date{}
\begin{document}

\maketitle

\begin{abstract}
There have been several efforts to extend distributional semantics beyond individual
words, to measure the similarity of word pairs, phrases, and sentences (briefly, {\em tuples};
ordered sets of words, contiguous or noncontiguous). One way to extend beyond words is to compare
two tuples using a function that combines pairwise similarities between the component words
in the tuples. A strength of this approach is that it works with both relational similarity (analogy)
and compositional similarity (paraphrase). However, past work required hand-coding the combination
function for different tasks. The main contribution of this paper is that combination
functions are generated by supervised learning. We achieve state-of-the-art results in measuring
relational similarity between word pairs (SAT analogies and Sem\-Eval~2012 Task~2) and measuring
compositional similarity between noun-modifier phrases and unigrams (multiple-choice paraphrase
questions).
\end{abstract}

\section{Introduction}

Harris~\shortcite{harris54} and Firth~\shortcite{firth57} hypothesized that words that appear
in similar contexts tend to have similar meanings. This hypothesis is the foundation for
distributional semantics, in which words are represented by context vectors. The similarity
of two words is calculated by comparing the two corresponding context vectors
\cite{lund95,landauer97,turney10}.

Distributional semantics is highly effective for measuring the semantic similarity between
individual words. On a set of eighty multiple-choice synonym questions from the test of English
as a foreign language (TOEFL), a distributional approach recently achieved 100\% accuracy
\cite{bullinaria12}. However, it has been difficult to extend distributional semantics beyond
individual words, to word pairs, phrases, and sentences.

Moving beyond individual words, there are various types of semantic similarity to consider.
Here we focus on paraphrase and analogy. Paraphrase is similarity in the meaning of two pieces
of text \cite{androutsopoulos10}. Analogy is similarity in the semantic relations of two sets
of words \cite{turney08b}.

It is common to study paraphrase at the sentence level \cite{androutsopoulos10}, but
we prefer to concentrate on the simplest type of paraphrase, where a bigram paraphrases a unigram.
For example, {\em dog house} is a paraphrase of {\em kennel}. In our experiments, we
concentrate on noun-modifier bigrams and noun unigrams.

Analogies map terms in one domain to terms in another domain \cite{gentner83}. The familiar
analogy between the solar system and the Rutherford-Bohr atomic model involves several terms
from the domain of the solar system and the domain of the atomic model \cite{turney08b}.

The simplest type of analogy is proportional analogy, which involves two pairs of words
\cite{turney06b}. For example, the pair {\em $\langle$cook, raw$\rangle$} is analogous to the
pair {\em $\langle$decorate, plain$\rangle$}. If we {\em cook} a thing, it is no longer {\em raw};
if we {\em decorate} a thing, it is no longer {\em plain}. The semantic relations between {\em cook}
and {\em raw} are similar to the semantic relations between {\em decorate} and {\em plain}.
In the following experiments, we focus on proportional analogies.

Erk~\shortcite{erk13} distinguished four approaches to extend distributional semantics beyond
words: In the first, a single vector space representation for a phrase or sentence is computed
from the representations of the individual words \cite{mitchell10,baroni10}. In the second,
two phrases or sentences are compared by combining multiple pairwise similarity values
\cite{socher11,turney12}. Third, weighted inference rules integrate distributional similarity
and formal logic \cite{garrette11}. Fourth, a single space integrates formal logic and vectors
\cite{clarke12}.

Taking the second approach, Turney~\shortcite{turney12} introduced a dual-space model, with
one space for measuring domain similarity (similarity of topic or field) and another for 
function similarity (similarity of role or usage). Similarities beyond individual words
are calculated by functions that combine domain and function similarities of component
words.

The dual-space model has been applied to measuring compositional similarity
(paraphrase recognition) and relational similarity (analogy recognition). In experiments
that tested for sensitivity to word order, the dual-space model performed significantly better
than competing approaches \cite{turney12}.

A limitation of past work with the dual-space model is that the combination functions were
hand-coded. Our main contribution is to show how hand-coding can be eliminated
with supervised learning. For ease of reference, we will call our approach {\em SuperSim}
(supervised similarity). With no modification of SuperSim for the specific task
(relational similarity or compositional similarity), we achieve better results than
previous hand-coded models.

Compositional similarity (paraphrase) compares two contiguous phrases or sentences
(\mbox{$n$-grams}), whereas relational similarity (analogy) does not require contiguity.
We use {\em tuple} to refer to both contiguous and noncontiguous word sequences.

We approach analogy as a problem of supervised tuple classification.
To measure the relational similarity between two word pairs, we train SuperSim with
quadruples that are labeled as positive and negative examples of analogies. For
example, the proportional analogy {\em $\langle$cook, raw, decorate, plain$\rangle$}
is labeled as a positive example.

A quadruple is represented by a feature vector, composed of domain and function similarities
from the dual-space model and other features based on corpus frequencies. SuperSim uses a support
vector machine \cite{platt98} to learn the probability that a quadruple $\langle a, b, c, d \rangle$
consists of a word pair $\langle a, b \rangle$ and an analogous word pair $\langle c, d \rangle$.
The probability can be interpreted as the degree of relational similarity between the two given
word pairs.

We also approach paraphrase as supervised tuple classification.
To measure the compositional similarity beween an \mbox{$m$-gram} and an \mbox{$n$-gram},
we train the learning algorithm with \mbox{$(m+n)$-tuples} that are positive and negative
examples of paraphrases.

SuperSim learns to estimate the probability that a triple $\langle a, b, c \rangle$
consists of a compositional bigram $ab$ and a synonymous unigram $c$. For instance,
the phrase {\em fish tank} is synonymous with {\em aquarium}; that is, {\em fish tank}
and {\em aquarium} have high compositional similarity. The triple
{\em $\langle$fish, tank, aquarium$\rangle$} is represented using the same 
features that we used for analogy. The probability of the triple can be interpreted
as the degree of compositional similarity between the given bigram and unigram.

We review related work in Section~\ref{sec:related}. The general feature space
for learning relations and compositions is presented in Section~\ref{sec:features}.
The experiments with relational similarity are described in Section~\ref{sec:relational},
and Section~\ref{sec:compositional} reports the results with compositional
similarity. Section~\ref{sec:discussion} discusses the implications of the results.
We consider future work in Section~\ref{sec:future} and conclude in Section~\ref{sec:conclusion}.

\section{Related Work}
\label{sec:related}

In SemEval~2012, Task~2 was concerned with measuring the degree of relational similarity
between two word pairs \cite{jurgens12} and Task~6 \cite{agirre12} examined the degree
of semantic equivalence between two sentences. These two areas of research have been
mostly independent, although Socher et al.~\shortcite{socher12} and Turney~\shortcite{turney12}
present unified perspectives on the two tasks. We first discuss some work on
relational similarity, then some work on compositional similarity, and lastly work
that unifies the two types of similarity.

\subsection{Relational Similarity}

LRA (latent relational analysis) measures relational similarity with a pair--pattern
matrix \cite{turney06b}. Rows in the matrix correspond to word pairs ($a, b$) and columns
correspond to patterns that connect the pairs (``$a$ for the $b$'') in a large corpus.
This is a {\em holistic} (noncompositional) approach to distributional similarity, since the
word pairs are opaque wholes; the component words have no separate representations.
A compositional approach to analogy has a representation for each word, and a word pair
is represented by composing the representations for each member of the pair. Given a vocabulary
of $N$ words, a compositional approach requires $N$ representations to handle all possible
word pairs, but a holistic approach requires $N^2$ representations. Holistic approaches do
not scale up \cite{turney12}. LRA required nine days to run.

Bollegala et al.~\shortcite{bollegala08} answered the SAT analogy questions with a support vector
machine trained on quadruples (proportional analogies), as we do here. However, their feature
vectors are holistic, and hence there are scaling problems.

Herda{\u{g}}delen and Baroni~\shortcite{herdagdelen09} used a support vector machine
to learn relational similarity. Their feature vectors contained a combination of holistic
and compositional features.

Measuring relational similarity is closely connected to classifying word pairs according
to their semantic relations \cite{turney05a}. Semantic relation classification was the
focus of SemEval~2007 Task~4 \cite{girju07} and SemEval~2010 Task~8 \cite{hendrickx10}.

\subsection{Compositional Similarity}

To extend distributional semantics beyond words, many researchers take the first approach
described by Erk~\shortcite{erk13}, in which a single vector space is used for individual
words, phrases, and sentences \cite{landauer97,mitchell08,mitchell10}. In this approach,
given the words $a$ and $b$ with context vectors $\mathbf{a}$ and $\mathbf{b}$,
we construct a vector for the bigram $ab$ by applying vector operations to $\mathbf{a}$
and $\mathbf{b}$.

Mitchell and Lapata~\shortcite{mitchell10} experiment with many different
vector operations and find that element-wise multiplication performs well. The
bigram $ab$ is represented by $\mathbf{c} = \mathbf{a} \odot \mathbf{b}$, where
$c_i = a_i \, \cdot \, b_i$. However, element-wise multiplication is commutative,
so the bigrams $ab$ and $ba$ map to the same vector $\mathbf{c}$. In experiments
that test for order sensitivity, element-wise multiplication performs poorly \cite{turney12}.

We can treat the bigram $ab$ as a unit, as if it were a single word, and construct
a context vector for $ab$ from occurrences of $ab$ in a large corpus. This holistic
approach to representing bigrams performs well when a limited set of bigrams is
specified in advance (before building the word--context matrix), but it does not scale up,
because there are too many possible bigrams \cite{turney12}.

Although the holistic approach does not scale up, we can generate a few holistic
bigram vectors and use them to train a supervised regression model \cite{guevara10,baroni10}.
Given a new bigram $cd$, not observed in the corpus, the regression model can predict
a holistic vector for $cd$, if $c$ and $d$ have been observed separately. We show in
Section~\ref{sec:compositional} that this idea can be adapted to train SuperSim without
manually labeled data.

Socher et al.~\shortcite{socher11} take the second approach described by Erk~\shortcite{erk13},
in which two sentences are compared by combining multiple pairwise similarity values.
They construct a variable-sized similarity matrix $\mathbf{X}$, in which the element $x_{ij}$
is the similarity between the \mbox{$i$-th} phrase of one sentence and the \mbox{$j$-th} phrase
of the other. Since supervised learning is simpler with fixed-sized feature vectors,
the variable-sized similarity matrix is then reduced to a smaller fixed-sized matrix, to
allow comparison of pairs of sentences of varying lengths.

\subsection{Unified Perspectives on Similarity}
\label{subsec:unified}

Socher et al.~\shortcite{socher12} represent words and phrases with a pair, consisting
of a vector and a matrix. The vector captures the meaning of the word or phrase
and the matrix captures how a word or phrase modifies the meaning of another word or
phrase when they are combined. They apply this matrix--vector representation to both
compositions and relations.

Turney~\shortcite{turney12} represents words with two vectors, a vector from domain
space and a vector from function space. The domain vector captures the topic or field
of the word and the function vector captures the functional role of the word. This dual-space
model is applied to both compositions and relations.

Here we extend the dual-space model of Turney~\shortcite{turney12} in two ways:
Hand-coding is replaced with supervised learning and two new sets of features
augment domain and function space. Moving to supervised learning instead of hand-coding
makes it easier to introduce new features.

In the dual-space model, parameterized similarity measures provided the input values for
hand-crafted functions. Each task required a different set of hand-crafted functions.
The parameters of the similarity measures were tuned using a customized grid search algorithm.
The grid search algorithm was not suitable for integration with a supervised learning algorithm.
The insight behind SuperSim is that, given appropriate features, a supervised learning
algorithm can replace the grid search algorithm and the hand-crafted functions.

\section{Features for Tuple Classification}
\label{sec:features}

We represent a tuple with four types of features, all based on
frequencies in a large corpus. The first type of feature is the logarithm
of the frequency of a word. The second type is the positive pointwise mutual information
(PPMI) between two words \cite{church89,bullinaria07}. Third and fourth are the
similarities of two words in domain and function space \cite{turney12}.

In the following experiments, we use the PPMI matrix from Turney et al.~\shortcite{turney11}
and the domain and function matrices from Turney~\shortcite{turney12}.\footnote{The three
matrices and the word frequency data are available on request from the author. The matrix 
files range from two to five gigabytes when packaged and compressed for distribution.} 
The three matrices and the word frequency data are based on
the same corpus, a collection of web pages gathered from university web sites, containing 
$5 \times 10^{10}$ words.\footnote{The corpus was collected by Charles Clarke at the University
of Waterloo. It is about 280 gigabytes of plain text.} All three matrices are word--context
matrices, in which the rows correspond to terms (words and phrases) in WordNet.\footnote{See
\url{http://wordnet.princeton.edu/} for information about WordNet.} The columns correspond
to the contexts in which the terms appear; each matrix involves a different kind of context.

Let $\langle x_1, x_2, \ldots, x_n \rangle$ be an \mbox{$n$-tuple} of words. The number
of features we use to represent this tuple increases as a function of $n$.

The first set of features consists of log frequency values for each word $x_i$ in the \mbox{$n$-tuple}.
Let $\textrm{freq}(x_i)$ be the frequency of $x_i$ in the corpus. We define $\textrm{LF}(x_i)$
as $\textrm{log}(\textrm{freq}(x_i) + 1)$. If $x_i$ is not in the corpus, $\textrm{freq}(x_i)$
is zero, and thus $\textrm{LF}(x_i)$ is also zero. There are $n$ log frequency features,
one  $\textrm{LF}(x_i)$ feature for each word in the \mbox{$n$-tuple}.

The second set of features consists of positive pointwise mutual information values
for each pair of words in the \mbox{$n$-tuple}. We use the raw PPMI matrix from Turney et
al.~\shortcite{turney11}. Although they computed the singular value decomposition (SVD)
to project the row vectors into a lower-dimensional space, we need the original high-dimensional
columns for our features. The raw PPMI matrix has 114,501 rows and 139,246 columns with a
density of 1.2\%. For each term in WordNet, there is a corresponding row in the raw PPMI matrix.
For each unigram in WordNet, there are two corresponding columns in the raw PPMI matrix, one
marked {\em left} and the other {\em right}.

Suppose $x_i$ corresponds to the \mbox{$i$-th} row of the PPMI matrix and $x_j$ corresponds
the \mbox{$j$-th} column, marked {\em left}. The value in the \mbox{$i$-th} row and \mbox{$j$-th}
column of the PPMI matrix, $\textrm{PPMI}(x_i, x_j, \textit{left})$, is the positive pointwise
mutual information of $x_i$ and $x_j$ co-occurring in the corpus, where $x_j$ is the first
word to the left of $x_i$, ignoring any intervening stop words (that is, ignoring any words
that are not in WordNet). If $x_i$ (or $x_j$) has no corresponding row (or column) in the matrix,
then the PPMI value is set to zero.

Turney et al.~\shortcite{turney11} estimated $\textrm{PPMI}(x_i, x_j, \textit{left})$
by sampling the corpus for phrases containing $x_i$ and then looking for $x_j$ to the
left of $x_i$ in the sampled phrases (and likewise for {\em right}). Due to this sampling
process, $\textrm{PPMI}(x_i, x_j, \textit{left})$ does not necessarily equal
$\textrm{PPMI}(x_j, x_i, \textit{right})$. For example, suppose $x_i$ is a rare
word and $x_j$ is a common word. With $\textrm{PPMI}(x_i, x_j, \textit{left})$, when
we sample phrases containing $x_i$, we are relatively likely to find $x_j$ in some of these
phrases. With $\textrm{PPMI}(x_j, x_i, \textit{right})$, when we sample phrases
containing $x_j$, we are less likely to find any phrases containing $x_i$. Although, in theory,
$\textrm{PPMI}(x_i, x_j, \textit{left})$ should equal $\textrm{PPMI}(x_j, x_i, \textit{right})$,
they are likely to be unequal given a limited sample.

From the \mbox{$n$-tuple}, we select all of the $n(n-1)$ pairs, $\langle x_i, x_j \rangle$, such
that $i \ne j$. We then generate two features for each pair, $\textrm{PPMI}(x_i, x_j, \textit{left})$
and $\textrm{PPMI}(x_i, x_j, \textit{right})$. Thus there are $2n(n-1)$ PPMI
values in the second set of features.

The third set of features consists of domain space similarity values for each pair of words
in the \mbox{$n$-tuple}. Domain space was designed to capture the topic of a word.
Turney~\shortcite{turney12} first constructed a frequency matrix, in which the rows correspond
to terms in WordNet and the columns correspond to nearby nouns. Given a term $x_i$, the corpus
was sampled for phrases containing $x_i$ and the phrases were processed with a part-of-speech
tagger, to identify nouns. If the noun $x_j$ was the closest noun
to the left or right of $x_i$, then the frequency count for the \mbox{$i$-th} row and
\mbox{$j$-th} column was incremented. The hypothesis was that the nouns near a term
characterize the topics associated with the term.

The word--context frequency matrix for domain space has 114,297 rows (terms) and 50,000 columns
(noun contexts, topics), with a density of 2.6\%. The frequency matrix was converted to a PPMI
matrix and then smoothed with SVD. The SVD yields three matrices, $\mathbf{U}$,
$\mathbf{\Sigma}$, and $\mathbf{V}$.

A term in domain space is represented by a row vector in $\mathbf{U}_k \mathbf{\Sigma}_k^p$.
The parameter $k$ specifies the number of singular values in the truncated singular
value decomposition; that is, $k$ is the number of latent factors in the low-dimensional
representation of the term \cite{landauer97}. We generate $\mathbf{U}_k$ and $\mathbf{\Sigma}_k$
by deleting the columns in $\mathbf{U}$ and $\mathbf{\Sigma}$ corresponding to the smallest
singular values. The parameter $p$ raises the singular values in $\mathbf{\Sigma}_k$ to the power
$p$ \cite{caron01}. As $p$ goes from one to zero, factors with smaller singular values are given
more weight. This has the effect of making the similarity measure more discriminating \cite{turney12}.

The similarity of two words in domain space, $\textrm{Dom}(x_i, x_j, k, p)$,
is computed by extracting the row vectors in $\mathbf{U}_k \mathbf{\Sigma}_k^p$
that correspond to the words $x_i$ and $x_j$, and then calculating their cosine.
Optimal performance requires tuning the parameters $k$ and $p$ for the task
\cite{bullinaria12,turney12}. In the following experiments,
we avoid directly tuning $k$ and $p$ by generating features with a variety of
values for $k$ and $p$, allowing the supervised learning
algorithm to decide which features to use.

From the \mbox{$n$-tuple}, we select all $\frac{1}{2} n(n-1)$ pairs, $\langle x_i, x_j \rangle$,
such that $i < j$. For each pair, we generate domain similarity features,
$\textrm{Dom}(x_i, x_j, k, p)$, where $k$ varies from 100 to 1000 in steps of 100 and
$p$ varies from 0 to 1 in steps of 0.1. The number of $k$ values, $n_k$, is 10 and the
number of $p$ values, $n_p$, is 11; therefore there are 110 features, $n_k n_p$, for
each pair, $\langle x_i, x_j \rangle$. Thus there are $\frac{1}{2} n(n-1) n_k n_p$
domain space similarity values in the third set of features.

The fourth set of features consists of function space similarity values for each pair of words
in the \mbox{$n$-tuple}. Function space was designed to capture the functional role of a word.
It is similar to domain space, except the context is based on verbal patterns,
instead of nearby nouns. The hypothesis was that the functional role of a word
is characterized by the patterns that relate the word to nearby verbs.

The word--context frequency matrix for function space has 114,101 rows (terms) and
50,000 columns (verb pattern contexts, functional roles), with a density of 1.2\%. The
frequency matrix was converted to a PPMI matrix and smoothed with SVD.

From the \mbox{$n$-tuple}, we select all $\frac{1}{2} n(n-1)$ pairs,
$\langle x_i, x_j \rangle$, such that $i < j$. For each pair, we generate function
similarity features, $\textrm{Fun}(x_i, x_j, k, p)$, where $k$ and $p$
vary as they did with domain space. Thus there are $\frac{1}{2} n(n-1) n_k n_p$
function space similarity values in the fourth set of features.

Table~\ref{tab:feats} summarizes the four sets of features and the size
of each set as a function of $n$, the number of words in the given tuple.
The values of $n_k$ and $n_p$ (10 and 11) are considered to be constants.
Table~\ref{tab:numfeats} shows the number of elements in the feature
vector, as $n$ varies from 1 to 6. The total number of features is $O(n^2)$.
We believe that this is acceptable growth and will scale up to comparing sentence pairs.

\begin{table}
\begin{center}
\scalebox{0.9}{
\begin{tabular}{ll}
\hline
Feature set                                       & Size of set \\
\hline
$\textrm{LF}(x_i)$                                & $n$ \\
$\textrm{PPMI}(x_i, x_j, \textit{handedness})$    & $2n(n-1)$ \\
$\textrm{Dom}(x_i, x_j, k, p)$                    & $\frac{1}{2} n(n-1) n_k n_p$ \\[2pt]
$\textrm{Fun}(x_i, x_j, k, p)$                    & $\frac{1}{2} n(n-1) n_k n_p$ \\[2pt]
\hline
\end{tabular}
} 
\end{center}
\caption {The four sets of features and their sizes.}
\label{tab:feats}
\end{table}

\begin{table}
\begin{center}
\scalebox{0.9}{
\begin{tabular}{ccrrrr}
\hline
\mbox{$n$-tuple} & LF  & PPMI & Dom  & Fun  & Total \\
\hline
      1          & 1   &  0   &    0 &    0 &    1 \\
      2          & 2   &  4   &  110 &  110 &  226 \\
      3          & 3   & 12   &  330 &  330 &  675 \\
      4          & 4   & 24   &  660 &  660 & 1348 \\
      5          & 5   & 40   & 1100 & 1100 & 2245 \\
      6          & 6   & 60   & 1650 & 1650 & 3366 \\
\hline
\end{tabular}
} 
\end{center}
\caption {Number of features for various tuple sizes.}
\label{tab:numfeats}
\end{table}

The four sets of features have a hierarchical relationship. The log frequency
features are based on counting isolated occurrences of each word in the corpus.
The PPMI features are based on direct co-occurrences of two words; that is, PPMI is only greater
than zero if the two words actually occur together in the corpus. Domain and function
space capture indirect or higher-order co-occurrence, due to the truncated SVD
\cite{lemaire06}; that is, the values of $\textrm{Dom}(x_i, x_j, k, p)$ and
$\textrm{Fun}(x_i, x_j, k, p)$ can be high even when $x_i$ and $x_j$
do not actually co-occur in the corpus. We conjecture that there are
yet higher orders in this hierarchy that would provide improved similarity
measures.

SuperSim learns to classify tuples by representing them with these features. SuperSim uses
the sequential minimal optimization (SMO) support vector machine (SVM) as implemented in Weka
\cite{platt98,witten11}.\footnote{Weka is available at \url{http://www.cs.waikato.ac.nz/ml/weka/}.}
The kernel is a normalized third-order polynomial. Weka provides probability estimates for
the classes by fitting the outputs of the SVM with logistic regression models.

\section{Relational Similarity}
\label{sec:relational}

This section presents experiments with learning relational similarity using SuperSim.
The training datasets consist of quadruples that are labeled as positive and negative
examples of analogies. Table~\ref{tab:numfeats} shows that the feature vectors
have 1,348 elements.

We experiment with three datasets, a collection of 374
five-choice questions from the SAT college entrance exam \cite{turney03b},
a modified ten-choice variation of the SAT questions \cite{turney12}, and
the relational similarity dataset from Sem\-Eval~2012 Task~2
\cite{jurgens12}.\footnote{The SAT questions are available on request from
the author. The Sem\-Eval~2012 Task~2 dataset is available at 
\url{https://sites.google.com/site/semeval2012task2/}.}

\subsection{Five-choice SAT Questions}

Table~\ref{tab:sat} is an example of a question from the 374 five-choice SAT questions.
Each five-choice question yields five labeled quadruples, by combining the stem with each
choice. The quadruple {\em $\langle$word, language, note, music$\rangle$} is labeled
{\em positive} and the other four quadruples are labeled {\em negative}.

\begin{table}
\begin{center}
\scalebox{0.9}{
\begin{tabular}{lll}
\hline
Stem:      &       & word:language \\
\hline
Choices:   & (1)   & paint:portrait \\
           & (2)   & poetry:rhythm \\
           & (3)   & note:music \\
           & (4)   & tale:story \\
           & (5)   & week:year \\
\hline
Solution:  & (3)   & note:music \\
\hline
\end{tabular}
} 
\end{center}
\caption {A five-choice SAT analogy question.}
\label{tab:sat}
\end{table}

Since learning works better with balanced training data \cite{japkowicz02},
we use the symmetries of proportional analogies to add more positive examples \cite{lepage96}.
For each positive quadruple, $\langle a, b, c, d \rangle$, we add three more positive
quadruples, $\langle b, a, d, c \rangle$, $\langle c, d, a, b \rangle$, and
$\langle d, c, b, a \rangle$. Thus each five-choice question provides
four positive and four negative quadruples.

We use ten-fold cross-validation to apply SuperSim to the SAT questions.
The folds are constructed so that the eight quadruples from each SAT question are
kept together in the same fold. To answer a question in the testing fold,
the learned model assigns a probability to each of the five choices and guesses
the choice with the highest probability. SuperSim achieves a score of 54.8\%
correct (205 out of 374).

Table~\ref{tab:topsat} gives the rank of SuperSim in the list of the top ten
results with the SAT analogy questions.\footnote{See the {\em State of the Art} page
on the ACL Wiki at \url{http://aclweb.org/aclwiki}.} The scores ranging from 51.1\%
to 57.0\% are not significantly different from SuperSim's score of 54.8\%, according to
Fisher's exact test at the 95\% confidence level. However, SuperSim answers
the SAT questions in a few minutes, whereas LRA requires nine days, and SuperSim
learns its models automatically, unlike the hand-coding of Turney~\shortcite{turney12}.

\begin{table}
\begin{center}
\scalebox{0.9}{
\begin{tabular}{llr}
\hline
Algorithm         & Reference                                              & Correct \\
\hline
Know-Best         & Veale~\shortcite{veale04}                              & 43.0 \\
k-means           & Bi{\c{c}}ici \& Yuret~\shortcite{bicici06}             & 44.0 \\
BagPack           & Herda{\u{g}}delen \& Baroni~\shortcite{herdagdelen09}  & 44.1 \\
VSM               & Turney \& Littman~\shortcite{turney05a}                & 47.1 \\
Dual-Space        & Turney~\shortcite{turney12}                            & 51.1 \\
BMI               & Bollegala et al.~\shortcite{bollegala09}               & 51.1 \\
PairClass         & Turney~\shortcite{turney08a}                           & 52.1 \\
PERT              & Turney~\shortcite{turney06a}                           & 53.5 \\
SuperSim          & ---                                                    & 54.8 \\
LRA               & Turney~\shortcite{turney06b}                           & 56.1 \\
Human             & Average college applicant                              & 57.0 \\
\hline
\end{tabular}
} 
\end{center}
\caption {The top ten results on five-choice SAT questions.}
\label{tab:topsat}
\end{table}

\subsection{Ten-choice SAT Questions}

In addition to symmetries, proportional analogies have asymmetries. In general, if
the quadruple $\langle a, b, c, d \rangle$ is positive, $\langle a, d, c, b \rangle$
is negative. For example, {\em $\langle$word, language, note, music$\rangle$}
is a good analogy, but {\em $\langle$word, music, note, language$\rangle$} is not.
{\em Words} are the basic units of {\em language} and {\em notes} are the basic units of
{\em music}, but {\em words} are not necessary for {\em music} and {\em notes} are not
necessary for {\em language}.

Turney~\shortcite{turney12} used this asymmetry to convert the 374 five-choice SAT
questions into 374 ten-choice SAT questions. Each choice $\langle c, d \rangle$ was
expanded with the stem $\langle a, b \rangle$, resulting in the quadruple
$\langle a, b, c, d \rangle$, and then the order was shuffled to $\langle a, d, c, b \rangle$,
so that each choice pair in a five-choice question generated two choice quadruples in a ten-choice
question. Nine of the quadruples are negative examples and the quadruple consisting of
the stem pair followed by the solution pair is the only positive example.
The purpose of the ten-choice questions is to test the ability of measures of relational
similarity to avoid the asymmetric distractors.

We use the ten-choice questions to compare the hand-coded dual-space approach
\cite{turney12} with SuperSim. We also use these questions to perform an
ablation study of the four sets of features in SuperSim. As
with the five-choice questions, we use the symmetries of proportional analogies to add
three more positive examples, so the training dataset has nine negative examples and
four positive examples per question. We apply ten-fold cross-validation to the 374 ten-choice
questions. 

On the ten-choice questions, SuperSim's score is 52.7\% (Table~\ref{tab:relfeats}),
compared to 54.8\% on the five-choice questions (Table~\ref{tab:topsat}), a drop
of 2.1\%. The hand-coded dual-space model scores 47.9\% (Table~\ref{tab:relfeats}),
compared to 51.1\% on the five-choice questions (Table~\ref{tab:topsat}), a drop
of 3.2\%. The difference between SuperSim (52.7\%) and the hand-coded dual-space model
(47.9\%) is not significant according to Fisher's exact test at the 95\% confidence level.
The advantage of SuperSim is that it does not need hand-coding. The results show
that SuperSim can avoid the asymmetric distractors.

\begin{table}
\begin{center}
\scalebox{0.9}{
\begin{tabular}{lccccr}
\hline
                  & \multicolumn{4}{c}{Features}             & \\
\cline{2-5}
Algorithm         & LF  & PPMI  & Dom  & Fun                 & Correct \\
\hline
Dual-Space        &  0  &  0    &  1   &  1                  & 47.9 \\
SuperSim          &  1  &  1    &  1   &  1                  & 52.7 \\
\hline
SuperSim          &  0  &  1    &  1   &  1                  & 52.7 \\
SuperSim          &  1  &  0    &  1   &  1                  & 52.7 \\
SuperSim          &  1  &  1    &  0   &  1                  & 45.7 \\
SuperSim          &  1  &  1    &  1   &  0                  & 41.7 \\
\hline
SuperSim          &  1  &  0    &  0   &  0                  &  5.6 \\
SuperSim          &  0  &  1    &  0   &  0                  & 32.4 \\
SuperSim          &  0  &  0    &  1   &  0                  & 39.6 \\
SuperSim          &  0  &  0    &  0   &  1                  & 39.3 \\
\hline
\end{tabular}
} 
\end{center}
\caption {Feature ablation with ten-choice SAT questions.}
\label{tab:relfeats}
\end{table}

Table~\ref{tab:relfeats} shows the impact of different subsets of features
on the percentage of correct answers to the ten-choice SAT questions. Included
features are marked 1 and ablated features are marked 0. The results show
that the log frequency (LF) and PPMI features are not helpful (but also
not harmful) for relational similarity. We also see that domain space and
function space are both needed for good results.

\subsection{SemEval~2012 Task~2}

The Sem\-Eval~2012 Task~2 dataset is based on the semantic relation classification
scheme of Bejar et al.~\shortcite{bejar91}, consisting of ten high-level
categories of relations and seventy-nine subcategories, with paradigmatic
examples of each subcategory. For instance, the subcategory {\em taxonomic}
in the category {\em class inclusion} has three paradigmatic examples,
{\em flower:tulip}, {\em emotion:rage}, and {\em poem:sonnet}. 

Jurgens et al.~\shortcite{jurgens12} used Amazon's Mechanical Turk to create
the Sem\-Eval~2012 Task~2 dataset in two phases. In the first phase, Turkers
expanded the paradigmatic examples for each subcategory to an average of forty-one
word pairs per subcategory, a total of 3,218 pairs. In the second phase, each word
pair from the first phase was assigned a prototypicality score, indicating its similarity
to the paradigmatic examples. The challenge of Sem\-Eval~2012 Task~2 was to 
guess the prototypicality scores.

SuperSim was trained on the five-choice SAT questions and evaluated on the SemEval~2012
Task~2 test dataset. For a given a word pair, we created quadruples, combining the word pair
with each of the paradigmatic examples for its subcategory. We then used SuperSim to
compute the probabilities for each quadruple. Our guess for the prototypicality
score of the given word pair was the average of the probabilities. Spearman's rank
correlation coefficient between the Turkers' prototypicality scores and SuperSim's
scores was 0.408, averaged over the sixty-nine subcategories in the testing set.
SuperSim has the highest Spearman correlation achieved to date on SemEval~2012 Task~2
(see Table~\ref{tab:topsemeval}).

\begin{table}
\begin{center}
\scalebox{0.9}{
\begin{tabular}{llr}
\hline
Algorithm         & Reference                              & Spearman \\
\hline
BUAP              & Tovar et al.~\shortcite{tovar12}       & 0.014 \\
Duluth-V2         & Pedersen~\shortcite{pedersen12}        & 0.038 \\
Duluth-V1         & Pedersen~\shortcite{pedersen12}        & 0.039 \\
Duluth-V0         & Pedersen~\shortcite{pedersen12}        & 0.050 \\
UTD-SVM           & Rink \& Harabagiu~\shortcite{rink12}   & 0.116 \\
UTD-NB            & Rink \& Harabagiu~\shortcite{rink12}   & 0.229 \\
RNN-1600          & Mikolov et al.~\shortcite{mikolov13}   & 0.275 \\
UTD-LDA           & Rink \& Harabagiu~\shortcite{rink13}   & 0.334 \\
Com               & Zhila et al.~\shortcite{zhila13}       & 0.353 \\
SuperSim          & ---                                    & 0.408 \\
\hline
\end{tabular}
} 
\end{center}
\caption {Spearman correlations for SemEval~2012 Task~2.}
\label{tab:topsemeval}
\end{table}

\section{Compositional Similarity}
\label{sec:compositional}

This section presents experiments using SuperSim to learn compositional similarity.
The datasets consist of triples, $\langle a, b, c \rangle$, such that $ab$ is a noun-modifier
bigram and $c$ is a noun unigram. The triples are labeled as positive and negative examples
of paraphrases. Table~\ref{tab:numfeats} shows that the feature vectors have 675 elements.
We experiment with two datasets, seven-choice and fourteen-choice noun-modifier questions
\cite{turney12}.\footnote{The seven-choice dataset is available at
\url{http://jair.org/papers/paper3640.html}. The fourteen-choice dataset can be
generated from the seven-choice dataset.}

\subsection{Noun-Modifier Questions}

The first dataset is a seven-choice noun-modifier question dataset, constructed from
WordNet \cite{turney12}. The dataset contains 680 questions for training and 1,500 for
testing, a total of 2,180 questions. Table~\ref{tab:fairyland} shows one of the
questions.

\begin{table}
\begin{center}
\scalebox{0.9}{
\begin{tabular}{lll}
\hline
Stem:      &       & fantasy world \\
\hline
Choices:   & (1)   & fairyland \\
           & (2)   & fantasy \\
           & (3)   & world \\
           & (4)   & phantasy \\
           & (5)   & universe \\
           & (6)   & ranter \\
           & (7)   & souring \\
\hline
Solution:  & (1)   & fairyland \\
\hline
\end{tabular}
} 
\end{center}
\caption {A noun-modifier question based on WordNet.}
\label{tab:fairyland}
\end{table}

The stem is a bigram and the choices are unigrams. The bigram is composed of a head
noun ({\em world}), modified by an adjective or noun ({\em fantasy}). The solution
is the unigram ({\em fairyland}) that belongs to the same WordNet synset as the stem.

The distractors are designed to be difficult for current approaches to composition.
For example, if {\em fantasy world} is represented by element-wise multiplication of
the context vectors for {\em fantasy} and {\em world} \cite{mitchell10}, the most
likely guess is {\em fantasy} or {\em world}, not {\em fairyland} \cite{turney12}.

Each seven-choice question yields seven labeled triples, by combining the stem
with each choice. The triple {\em $\langle$fantasy, world, fairyland$\rangle$} is labeled
{\em positive} and the other six triples are labeled {\em negative}.

In general, if $\langle a, b, c \rangle$ is a positive example, then $\langle b, a, c \rangle$
is negative. For example, {\em world fantasy} is not a paraphrase of {\em fairyland}. The second
dataset is constructed by applying this shuffling transformation to convert the 2,180
seven-choice questions into 2,180 fourteen-choice questions \cite{turney12}. The second dataset
is designed to be difficult for approaches that are not sensitive to word order.

Table~\ref{tab:nounmod} shows the percentage of the testing questions that are
answered correctly for the two datasets. Because vector addition and element-wise
multiplication are not sensitive to word order, they perform poorly on the
fourteen-choice questions. For both datasets, SuperSim performs significantly better than all
other approaches, except for the holistic approach, according to Fisher's exact test at the
95\% confidence level.\footnote{The results for SuperSim are new but the other results
in Table~\ref{tab:nounmod} are from Turney~\shortcite{turney12}.}

\begin{table}
\begin{center}
\scalebox{0.9}{
\begin{tabular}{lrr}
\hline
                               & \multicolumn{2}{c}{Correct} \\
\cline{2-3}
Algorithm                      & 7-choices  & 14-choices \\
\hline
Vector addition                & 50.1       & 22.5 \\
Element-wise multiplication    & 57.5       & 27.4 \\
Dual-Space model               & 58.3       & 41.5 \\
SuperSim                       & 75.9       & 68.0 \\
Holistic model                 & 81.6       & --- \\
\hline
\end{tabular}
} 
\end{center}
\caption {Results for the two noun-modifier datasets.}
\label{tab:nounmod}
\end{table}

The holistic approach is noncompositional. The stem bigram is represented by a
single context vector, generated by treating the bigram as if it were a unigram. A
noncompositional approach cannot scale up to realistic applications \cite{turney12}.
The holistic approach cannot be applied to the fourteen-choice questions, because the
bigrams in these questions do not correspond to terms in WordNet, and hence
they do not correspond to row vectors in the matrices we use (see
Section~\ref{sec:features}).

Turney~\shortcite{turney12} found it necessary to hand-code a soundness check
into all of the algorithms (vector addition, element-wise multiplication, dual-space,
and holistic). Given a stem $ab$ and a choice $c$, the hand-coded check assigns a
minimal score to the choice if $c = a$ or $c = b$. We do not need to hand-code any
checking into SuperSim. It learns automatically from the training data to avoid
such choices.

\subsection{Ablation Experiments}

Table~\ref{tab:nmodfeats} shows the effects of ablating sets of features on the
performance of SuperSim with the fourteen-choice questions. PPMI features are
the most important; by themselves, they achieve 59.7\% correct, although the other
features are needed to reach 68.0\%. Domain space features reach the second highest
performance when used alone (34.6\%), but they reduce performance (from 69.3\% to 68.0\%)
when combined with other features; however, the drop is not significant according to
Fisher's exact test at the 95\% significance level.

\begin{table}
\begin{center}
\scalebox{0.9}{
\begin{tabular}{lccccr}
\hline
                  & \multicolumn{4}{c}{Features}             & \\
\cline{2-5}
Algorithm         & LF  & PPMI  & Dom  & Fun                 & Correct \\
\hline
Dual-Space        &  0  &  0    &  1   &  1                  & 41.5 \\
SuperSim          &  1  &  1    &  1   &  1                  & 68.0 \\
\hline
SuperSim          &  0  &  1    &  1   &  1                  & 66.6 \\
SuperSim          &  1  &  0    &  1   &  1                  & 52.3 \\
SuperSim          &  1  &  1    &  0   &  1                  & 69.3 \\
SuperSim          &  1  &  1    &  1   &  0                  & 65.9 \\
\hline
SuperSim          &  1  &  0    &  0   &  0                  & 14.1 \\
SuperSim          &  0  &  1    &  0   &  0                  & 59.7 \\
SuperSim          &  0  &  0    &  1   &  0                  & 34.6 \\
SuperSim          &  0  &  0    &  0   &  1                  & 32.9 \\
\hline
\end{tabular}
} 
\end{center}
\caption {Ablation with fourteen-choice questions.}
\label{tab:nmodfeats}
\end{table}

Since the PPMI features play an important role in answering the noun-modifier
questions, let us take a closer look at them. From Table~\ref{tab:numfeats},
we see that there are twelve PPMI features for the triple $\langle a, b, c \rangle$,
where $ab$ is a noun-modifier bigram and $c$ is a noun unigram.
We can split the twelve features into three subsets, one subset for each pair
of words, $\langle a, b \rangle$, $\langle a, c \rangle$, and $\langle b, c \rangle$.
For example, the subset for $\langle a, b \rangle$ is the four features
$\textrm{PPMI}(a, b, \textit{left})$, $\textrm{PPMI}(b, a, \textit{left})$,
$\textrm{PPMI}(a, b, \textit{right})$, and $\textrm{PPMI}(b, a, \textit{right})$.
Table~\ref{tab:ppmifeats} shows the effects of ablating these subsets.

\begin{table}
\begin{center}
\scalebox{0.9}{
\begin{tabular}{cccr}
\hline
\multicolumn{3}{c}{PPMI feature subsets}                                 & \\[1pt]
\cline{1-3}
$\langle a, b \rangle$ & $\langle a, c \rangle$ & $\langle b, c \rangle$ & Correct \\
\hline
      1                &        1               &        1               & 68.0 \\
\hline
      0                &        1               &        1               & 59.9 \\
      1                &        0               &        1               & 65.4 \\
      1                &        1               &        0               & 67.5 \\
\hline
      1                &        0               &        0               & 62.6 \\
      0                &        1               &        0               & 58.1 \\
      0                &        0               &        1               & 55.6 \\
\hline
      0                &        0               &        0               & 52.3 \\
\hline
\end{tabular}
} 
\end{center}
\caption {PPMI subset ablation with fourteen-choices.}
\label{tab:ppmifeats}
\end{table}

The results in Table~\ref{tab:ppmifeats} indicate that all three PPMI subsets
contribute to the performance of SuperSim, but the $\langle a, b \rangle$
subset contributes more than the other two subsets. The $\langle a, b \rangle$
features help to increase the sensitivity of SuperSim to the order of the words
in the noun-modifier bigram; for example, they make it easier to distinguish
{\em fantasy world} from {\em world fantasy}.

\subsection{Holistic Training}

SuperSim uses 680 training questions to learn to recognize when a bigram is
a paraphrase of a unigram; it learns from expert knowledge implicit in
WordNet synsets. It would be advantageous to be able to train SuperSim with
less reliance on expert knowledge.

Past work with adjective-noun bigrams has shown that we can use holistic bigram
vectors to train a supervised regression model \cite{guevara10,baroni10}. The
output of the regression model is a vector representation for a bigram that
approximates the holistic vector for the bigram; that is, it approximates
the vector we would get by treating the bigram as if it were a unigram.

SuperSim does not generate vectors as output, but we can still use holistic
bigram vectors for training. Table~\ref{tab:search} shows a seven-choice training
question that was generated without using WordNet synsets. The choices of
the form $a\_b$ are bigrams, but we represent them with holistic bigram vectors;
we pretend they are unigrams. We call $a\_b$ bigrams {\em pseudo-unigrams}.
As far as SuperSim is concerned, there is no difference between these pseudo-unigrams
and true unigrams. The question in Table~\ref{tab:search} is treated
the same as the question in Table~\ref{tab:fairyland}.

\begin{table}
\begin{center}
\scalebox{0.9}{
\begin{tabular}{lll}
\hline
Stem:      &       & search engine \\
\hline
Choices:   & (1)   & search\_engine \\
           & (2)   & search \\
           & (3)   & engine \\
           & (4)   & search\_language \\
           & (5)   & search\_warrant \\
           & (6)   & diesel\_engine \\
           & (7)   & steam\_engine \\
\hline
Solution:  & (1)   & search\_engine \\
\hline
\end{tabular}
} 
\end{center}
\caption {A question based on holistic vectors.}
\label{tab:search}
\end{table}

We generate 680 holistic training questions by randomly selecting 680 noun-modifier
bigrams from WordNet as stems for the questions ({\em search engine}), avoiding any
bigrams that appear as stems in the testing questions. The solution ({\em search\_engine})
is the pseudo-unigram that corresponds to the stem bigram. In the matrices in
Section~\ref{sec:features}, each term in WordNet corresponds to a row vector. These
corresponding row vectors enable us to treat bigrams from WordNet as if they were unigrams.
The distractors are the component unigrams in the stem bigram ({\em search} and {\em engine})
and pseudo-unigrams that share a component word with the stem ({\em search\_warrant},
{\em diesel\_engine}). To construct the holistic training questions, we used
WordNet as a source of bigrams, but we ignored the rich information that WordNet provides
about these bigrams, such as their synonyms, hypernyms, hyponyms, meronyms, and glosses.

Table~\ref{tab:holistic} compares holistic training to standard training (that is,
training with questions like Table~\ref{tab:search} versus training with questions
like Table~\ref{tab:fairyland}). The testing set is the standard testing set
in both cases. There is a significant drop in performance with holistic training,
but the performance still surpasses vector addition, element-wise multiplication,
and the hand-coded dual-space model (see Table~\ref{tab:nounmod}).

\begin{table}
\begin{center}
\scalebox{0.9}{
\begin{tabular}{lrr}
\hline
            & \multicolumn{2}{c}{Correct} \\
\cline{2-3}
Training    & 7-choices  & 14-choices \\
\hline
Holistic    & 61.8       & 54.4 \\
Standard    & 75.9       & 68.0 \\
\hline
\end{tabular}
} 
\end{center}
\caption {Results for SuperSim with holistic training.}
\label{tab:holistic}
\end{table}

Since holistic questions can be generated automatically without human expertise,
we experimented with increasing the size of the holistic training dataset,
growing it from 1,000 to 10,000 questions in increments of 1,000. The performance
on the fourteen-choice questions with holistic training and standard testing
varied between 53.3\% and 55.1\% correct, with no clear trend up or down. This is not
significantly different from the performance with 680 holistic training questions (54.4\%).

It seems likely that the drop in performance with holistic training instead
of standard training is due to a difference in the nature of the standard
questions (Table~\ref{tab:fairyland}) and the holistic questions (Table~\ref{tab:search}).
We are currently investigating this issue. We expect to be able to close the
performance gap in future work, by improving the holistic questions. However,
it is possible that there are fundamental limits to holistic training.

\section{Discussion}
\label{sec:discussion}

SuperSim performs slightly better (not statistically significant) than the hand-coded
dual-space model on relational similarity problems (Section~\ref{sec:relational}), but
it performs much better on compositional similarity problems (Section~\ref{sec:compositional}).
The ablation studies suggest this is due to the PPMI features, which have no effect on
ten-choice SAT performance (Table~\ref{tab:relfeats}), but have a large
effect on fourteen-choice noun-modifier paraphrase performance (Table~\ref{tab:nmodfeats}).

One advantage of supervised learning over hand-coding is that it facilitates adding
new features. It is not clear how to modify the hand-coded equations for the dual-space
model of noun-modifier composition \cite{turney12} to include PPMI information.

SuperSim is one of the few approaches to distributional semantics beyond words
that has attempted to address both relational and compositional similarity
(see Section~\ref{subsec:unified}). It is a strength of this approach that it
works well with both kinds of similarity. 

\section{Future Work and Limitations}
\label{sec:future}

Given the promising results with holistic training for noun-modifier paraphrases,
we plan to experiment with holistic training for analogies. Consider the proportional
analogy {\em hard} is to {\em hard\_time} as {\em good} is to {\em good\_time},
where {\em hard\_time} and {\em good\_time} are pseudo-unigrams. To a human, this
analogy is trivial, but SuperSim has no access to the surface form of a term.
As far as SuperSim is concerned, this analogy is much the same as the analogy
{\em hard} is to {\em difficulty} as {\em good} is to {\em fun}. This strategy
automatically converts simple, easily generated analogies into more complex,
challenging analogies, which may be suited to training SuperSim.

This also suggests that noun-modifier paraphrases may be used to solve analogies.
Perhaps we can evaluate the quality of a candidate analogy $\langle a, b, c, d \rangle$
by searching for a term $e$ such that $\langle b, e, a \rangle$ and $\langle d, e, c \rangle$
are good paraphrases. For example, consider the analogy {\em mason} is to {\em stone}
as {\em carpenter} is to {\em wood}. We can paraphrase {\em mason} as {\em stone worker}
and {\em carpenter} as {\em wood worker}. This transforms the analogy to
{\em stone worker} is to {\em stone} as {\em wood worker} is to {\em wood},
which makes it easier to recognize the relational similarity.

Another area for future work is extending SuperSim beyond noun-modifier paraphrases
to measuring the similarity of sentence pairs. We plan to adapt ideas from
Socher et al.~\shortcite{socher11} for this task. They use {\em dynamic pooling}
to represent sentences of varying size with fixed-size feature vectors. 
Using fixed-size feature vectors avoids the problem of quadratic growth and
it enables the supervised learner to generalize over sentences of varying length.

Some of the competing approaches discussed by Erk~\shortcite{erk13} incorporate
formal logic. The work of Baroni et al.~\shortcite{baroni12} suggests ways that
SuperSim could be developed to deal with logic.

We believe that SuperSim could benefit from more features, with greater diversity.
One place to look for these features is higher levels in the hierarchy that
we sketch in Section~\ref{sec:features}.

Our ablation experiments suggest that domain and function spaces provide the most
important features for relational similarity, but PPMI values provide the most
important features for noun-modifier compositional similarity. Explaining this is
another topic for future research.

\section{Conclusion}
\label{sec:conclusion}

In this paper, we have presented SuperSim, a unified approach to analogy (relational similarity)
and paraphrase (compositional similarity). SuperSim treats them both as problems of supervised
tuple classification. The supervised learning algorithm is a standard support vector machine.
The main contribution of SuperSim is a set of four types of features for representing tuples.
The features work well with both analogy and paraphrase, with no task-specific modifications.
SuperSim matches the state of the art on SAT analogy questions and substantially advances the state
of the art on the SemEval~2012 Task~2 challenge and the noun-modifier paraphrase questions.

SuperSim runs much faster than LRA \cite{turney06b}, answering the SAT questions in minutes
instead of days. Unlike the dual-space model \cite{turney12}, SuperSim requires no hand-coded
similarity composition functions. Since there is no hand-coding, it is easy to add new features
to SuperSim. Much work remains to be done, such as incorporating logic and scaling up
to sentence paraphrases, but past work suggests that these problems are tractable.

In the four approaches described by Erk~\shortcite{erk13}, SuperSim is an instance of the
second approach to extending distributional semantics beyond words, comparing word
pairs, phrases, or sentences (in general, tuples) by combining multiple pairwise similarity
values. Perhaps the main significance of this paper is that it provides some evidence in
support of this general approach.

\bibliographystyle{acl2012}
\bibliography{learning-similarity}

\begin{thebibliography}{}

\bibitem[\protect\citename{Agirre \bgroup et al.\egroup }2012]{agirre12}
Eneko Agirre, Daniel Cer, Mona Diab, and Aitor Gonzalez-Agirre.
\newblock 2012.
\newblock {Semeval-2012 Task 6}: A pilot on semantic textual similarity.
\newblock In {\em Proceedings of the First Joint Conference on Lexical and
  Computational Semantics (*SEM)}, pages 385--393, Montr{\'e}al, Canada.

\bibitem[\protect\citename{Androutsopoulos and
  Malakasiotis}2010]{androutsopoulos10}
Ion Androutsopoulos and Prodromos Malakasiotis.
\newblock 2010.
\newblock A survey of paraphrasing and textual entailment methods.
\newblock {\em Journal of Artificial Intelligence Research}, 38:135--187.

\bibitem[\protect\citename{Baroni and Zamparelli}2010]{baroni10}
Marco Baroni and Roberto Zamparelli.
\newblock 2010.
\newblock Nouns are vectors, adjectives are matrices: Representing
  adjective-noun constructions in semantic space.
\newblock In {\em Proceedings of the 2010 Conference on Empirical Methods in
  Natural Language Processing (EMNLP 2010)}, pages 1183--1193.

\bibitem[\protect\citename{Baroni \bgroup et al.\egroup }2012]{baroni12}
Marco Baroni, Raffaella Bernardi, Ngoc-Quynh Do, and {Chung-chieh} Shan.
\newblock 2012.
\newblock Entailment above the word level in distributional semantics.
\newblock In {\em Proceedings of the 13th Conference of the European Chapter of
  the Association for Computational Linguistics (EACL 2012)}, pages 23--32.

\bibitem[\protect\citename{Bejar \bgroup et al.\egroup }1991]{bejar91}
Isaac~I. Bejar, Roger Chaffin, and Susan~E. Embretson.
\newblock 1991.
\newblock {\em Cognitive and Psychometric Analysis of Analogical Problem
  Solving}.
\newblock Springer-Verlag.

\bibitem[\protect\citename{Bi{\c{c}}ici and Yuret}2006]{bicici06}
Ergun Bi{\c{c}}ici and Deniz Yuret.
\newblock 2006.
\newblock Clustering word pairs to answer analogy questions.
\newblock In {\em Proceedings of the Fifteenth Turkish Symposium on Artificial
  Intelligence and Neural Networks (TAINN 2006)}, Akyaka, Mugla, Turkey.

\bibitem[\protect\citename{Bollegala \bgroup et al.\egroup }2008]{bollegala08}
Danushka Bollegala, Yutaka Matsuo, and Mitsuru Ishizuka.
\newblock 2008.
\newblock {WWW} sits the {SAT}: Measuring relational similarity on the {Web}.
\newblock In {\em Proceedings of the 18th European Conference on Artificial
  Intelligence (ECAI 2008)}, pages 333--337, Patras, Greece.

\bibitem[\protect\citename{Bollegala \bgroup et al.\egroup }2009]{bollegala09}
Danushka Bollegala, Yutaka Matsuo, and Mitsuru Ishizuka.
\newblock 2009.
\newblock Measuring the similarity between implicit semantic relations from the
  {Web}.
\newblock In {\em Proceedings of the 18th International Conference on World
  Wide Web (WWW 2009)}, pages 651--660.

\bibitem[\protect\citename{Bullinaria and Levy}2007]{bullinaria07}
John Bullinaria and Joseph Levy.
\newblock 2007.
\newblock Extracting semantic representations from word co-occurrence
  statistics: A computational study.
\newblock {\em Behavior Research Methods}, 39(3):510--526.

\bibitem[\protect\citename{Bullinaria and Levy}2012]{bullinaria12}
John Bullinaria and Joseph Levy.
\newblock 2012.
\newblock Extracting semantic representations from word co-occurrence
  statistics: Stop-lists, stemming, and {SVD}.
\newblock {\em Behavior Research Methods}, 44(3):890--907.

\bibitem[\protect\citename{Caron}2001]{caron01}
John Caron.
\newblock 2001.
\newblock Experiments with {LSA} scoring: Optimal rank and basis.
\newblock In {\em Proceedings of the SIAM Computational Information Retrieval
  Workshop}, pages 157--169, Raleigh, NC.

\bibitem[\protect\citename{Church and Hanks}1989]{church89}
Kenneth Church and Patrick Hanks.
\newblock 1989.
\newblock Word association norms, mutual information, and lexicography.
\newblock In {\em Proceedings of the 27th Annual Conference of the Association
  of Computational Linguistics}, pages 76--83, Vancouver, British Columbia.

\bibitem[\protect\citename{Clarke}2012]{clarke12}
Daoud Clarke.
\newblock 2012.
\newblock A context-theoretic framework for compositionality in distributional
  semantics.
\newblock {\em Computational Linguistics}, 38(1):41--71.

\bibitem[\protect\citename{Erk}2013]{erk13}
Katrin Erk.
\newblock 2013.
\newblock Towards a semantics for distributional representations.
\newblock In {\em Proceedings of the 10th International Conference on
  Computational Semantics (IWCS 2013)}, Potsdam, Germany.

\bibitem[\protect\citename{Firth}1957]{firth57}
John~Rupert Firth.
\newblock 1957.
\newblock A synopsis of linguistic theory 1930--1955.
\newblock In {\em Studies in Linguistic Analysis}, pages 1--32. Blackwell,
  Oxford.

\bibitem[\protect\citename{Garrette \bgroup et al.\egroup }2011]{garrette11}
Dan Garrette, Katrin Erk, and Ray Mooney.
\newblock 2011.
\newblock Integrating logical representations with probabilistic information
  using markov logic.
\newblock In {\em Proceedings of the 9th International Conference on
  Computational Semantics (IWCS 2011)}, pages 105--114.

\bibitem[\protect\citename{Gentner}1983]{gentner83}
Dedre Gentner.
\newblock 1983.
\newblock Structure-mapping: A theoretical framework for analogy.
\newblock {\em Cognitive Science}, 7(2):155--170.

\bibitem[\protect\citename{Girju \bgroup et al.\egroup }2007]{girju07}
Roxana Girju, Preslav Nakov, Vivi Nastase, Stan Szpakowicz, Peter Turney, and
  Deniz Yuret.
\newblock 2007.
\newblock Semeval-2007 task 04: Classification of semantic relations between
  nominals.
\newblock In {\em Proceedings of the Fourth International Workshop on Semantic
  Evaluations (SemEval 2007)}, pages 13--18, Prague, Czech Republic.

\bibitem[\protect\citename{Guevara}2010]{guevara10}
Emiliano Guevara.
\newblock 2010.
\newblock A regression model of adjective-noun compositionality in
  distributional semantics.
\newblock In {\em Proceedings of the 2010 Workshop on GEometrical Models of
  Natural Language Semantics (GEMS 2010)}, pages 33--37.

\bibitem[\protect\citename{Harris}1954]{harris54}
Zellig Harris.
\newblock 1954.
\newblock Distributional structure.
\newblock {\em Word}, 10(23):146--162.

\bibitem[\protect\citename{Hendrickx \bgroup et al.\egroup }2010]{hendrickx10}
Iris Hendrickx, Su~Nam Kim, Zornitsa Kozareva, Preslav Nakov, Diarmuid~\'{O}
  S\'{e}aghdha, Sebastian Pad\'{o}, Marco Pennacchiotti, Lorenza Romano, and
  Stan Szpakowicz.
\newblock 2010.
\newblock Semeval-2010 task 8: Multi-way classification of semantic relations
  between pairs of nominals.
\newblock In {\em Proceedings of the 5th International Workshop on Semantic
  Evaluation}, pages 33--38, Uppsala, Sweden.

\bibitem[\protect\citename{Herda{\u{g}}delen and Baroni}2009]{herdagdelen09}
Ama{\c{c}} Herda{\u{g}}delen and Marco Baroni.
\newblock 2009.
\newblock Bagpack: A general framework to represent semantic relations.
\newblock In {\em Proceedings of the EACL 2009 Geometrical Models for Natural
  Language Semantics (GEMS) Workshop}, pages 33--40.

\bibitem[\protect\citename{Japkowicz and Stephen}2002]{japkowicz02}
Nathalie Japkowicz and Shaju Stephen.
\newblock 2002.
\newblock The class imbalance problem: A systematic study.
\newblock {\em Intelligent Data Analysis}, 6(5):429--449.

\bibitem[\protect\citename{Jurgens \bgroup et al.\egroup }2012]{jurgens12}
David~A. Jurgens, Saif~M. Mohammad, Peter~D. Turney, and Keith~J. Holyoak.
\newblock 2012.
\newblock {SemEval-2012 Task 2}: Measuring degrees of relational similarity.
\newblock In {\em Proceedings of the First Joint Conference on Lexical and
  Computational Semantics (*SEM)}, pages 356--364, Montr{\'e}al, Canada.

\bibitem[\protect\citename{Landauer and Dumais}1997]{landauer97}
Thomas~K. Landauer and Susan~T. Dumais.
\newblock 1997.
\newblock A solution to {Plato's} problem: The latent semantic analysis theory
  of the acquisition, induction, and representation of knowledge.
\newblock {\em Psychological Review}, 104(2):211--240.

\bibitem[\protect\citename{Lemaire and Denhi{\`e}re}2006]{lemaire06}
Beno{\^i}t Lemaire and Guy Denhi{\`e}re.
\newblock 2006.
\newblock Effects of high-order co-occurrences on word semantic similarity.
\newblock {\em Current Psychology Letters: Behaviour, Brain \& Cognition},
  18(1).

\bibitem[\protect\citename{Lepage and Shin-ichi}1996]{lepage96}
Yves Lepage and Ando Shin-ichi.
\newblock 1996.
\newblock Saussurian analogy: A theoretical account and its application.
\newblock In {\em Proceedings of the 16th International Conference on
  Computational Linguistics (COLING 1996)}, pages 717--722.

\bibitem[\protect\citename{Lund \bgroup et al.\egroup }1995]{lund95}
Kevin Lund, Curt Burgess, and Ruth~Ann Atchley.
\newblock 1995.
\newblock Semantic and associative priming in high-dimensional semantic space.
\newblock In {\em Proceedings of the 17th Annual Conference of the Cognitive
  Science Society}, pages 660--665.

\bibitem[\protect\citename{Mikolov \bgroup et al.\egroup }2013]{mikolov13}
Tomas Mikolov, {Wen-tau} Yih, and Geoffrey Zweig.
\newblock 2013.
\newblock Linguistic regularities in continuous space word representations.
\newblock In {\em Proceedings of the 2013 Conference of the North American
  Chapter of the Association for Computational Linguistics: Human Language
  Technologies (NAACL 2013)}, Atlanta, Georgia.

\bibitem[\protect\citename{Mitchell and Lapata}2008]{mitchell08}
Jeff Mitchell and Mirella Lapata.
\newblock 2008.
\newblock Vector-based models of semantic composition.
\newblock In {\em Proceedings of ACL-08: HLT}, pages 236--244, Columbus, Ohio.
  Association for Computational Linguistics.

\bibitem[\protect\citename{Mitchell and Lapata}2010]{mitchell10}
Jeff Mitchell and Mirella Lapata.
\newblock 2010.
\newblock Composition in distributional models of semantics.
\newblock {\em Cognitive Science}, 34(8):1388--1429.

\bibitem[\protect\citename{Pedersen}2012]{pedersen12}
Ted Pedersen.
\newblock 2012.
\newblock Duluth: Measuring degrees of relational similarity with the gloss
  vector measure of semantic relatedness.
\newblock In {\em First Joint Conference on Lexical and Computational Semantics
  (*SEM)}, pages 497--501, Montreal, Canada.

\bibitem[\protect\citename{Platt}1998]{platt98}
John~C. Platt.
\newblock 1998.
\newblock Fast training of support vector machines using sequential minimal
  optimization.
\newblock In {\em Advances in Kernel Methods: Support Vector Learning}, pages
  185--208, Cambridge, MA. MIT Press.

\bibitem[\protect\citename{Rink and Harabagiu}2012]{rink12}
Bryan Rink and Sanda Harabagiu.
\newblock 2012.
\newblock {UTD}: Determining relational similarity using lexical patterns.
\newblock In {\em First Joint Conference on Lexical and Computational Semantics
  (*SEM)}, pages 413--418, Montreal, Canada.

\bibitem[\protect\citename{Rink and Harabagiu}2013]{rink13}
Bryan Rink and Sanda Harabagiu.
\newblock 2013.
\newblock The impact of selectional preference agreement on semantic relational
  similarity.
\newblock In {\em Proceedings of the 10th International Conference on
  Computational Semantics (IWCS 2013)}, Potsdam, Germany.

\bibitem[\protect\citename{Socher \bgroup et al.\egroup }2011]{socher11}
Richard Socher, Eric~H. Huang, Jeffrey Pennington, Andrew~Y. Ng, and
  Christopher~D. Manning.
\newblock 2011.
\newblock Dynamic pooling and unfolding recursive autoencoders for paraphrase
  detection.
\newblock In {\em Advances in Neural Information Processing Systems (NIPS
  2011)}, pages 801--809.

\bibitem[\protect\citename{Socher \bgroup et al.\egroup }2012]{socher12}
Richard Socher, Brody Huval, Christopher Manning, and Andrew Ng.
\newblock 2012.
\newblock Semantic compositionality through recursive matrix-vector spaces.
\newblock In {\em Proceedings of the 2012 Joint Conference on Empirical Methods
  in Natural Language Processing and Computational Natural Language Learning
  (EMNLP-CoNLL 2012)}, pages 1201--1211.

\bibitem[\protect\citename{Tovar \bgroup et al.\egroup }2012]{tovar12}
Mireya Tovar, J.~Alejandro Reyes, Azucena Montes, Darnes Vilari{\~n}o, David
  Pinto, and Saul Le{\'o}n.
\newblock 2012.
\newblock {BUAP}: A first approximation to relational similarity measuring.
\newblock In {\em First Joint Conference on Lexical and Computational Semantics
  (*SEM)}, pages 502--505, Montreal, Canada.

\bibitem[\protect\citename{Turney and Littman}2005]{turney05a}
Peter~D. Turney and Michael~L. Littman.
\newblock 2005.
\newblock Corpus-based learning of analogies and semantic relations.
\newblock {\em Machine Learning}, 60(1--3):251--278.

\bibitem[\protect\citename{Turney and Pantel}2010]{turney10}
Peter~D. Turney and Patrick Pantel.
\newblock 2010.
\newblock From frequency to meaning: Vector space models of semantics.
\newblock {\em Journal of Artificial Intelligence Research}, 37:141--188.

\bibitem[\protect\citename{Turney \bgroup et al.\egroup }2003]{turney03b}
Peter~D. Turney, Michael~L. Littman, Jeffrey Bigham, and Victor Shnayder.
\newblock 2003.
\newblock Combining independent modules to solve multiple-choice synonym and
  analogy problems.
\newblock In {\em Proceedings of the International Conference on Recent
  Advances in Natural Language Processing (RANLP-03)}, pages 482--489,
  Borovets, Bulgaria.

\bibitem[\protect\citename{Turney \bgroup et al.\egroup }2011]{turney11}
Peter~D. Turney, Yair Neuman, Dan Assaf, and Yohai Cohen.
\newblock 2011.
\newblock Literal and metaphorical sense identification through concrete and
  abstract context.
\newblock In {\em Proceedings of the 2011 Conference on Empirical Methods in
  Natural Language Processing}, pages 680--690.

\bibitem[\protect\citename{Turney}2006a]{turney06a}
Peter~D. Turney.
\newblock 2006a.
\newblock Expressing implicit semantic relations without supervision.
\newblock In {\em Proceedings of the 21st International Conference on
  Computational Linguistics and 44th Annual Meeting of the Association for
  Computational Linguistics (Coling/ACL-06)}, pages 313--320, Sydney,
  Australia.

\bibitem[\protect\citename{Turney}2006b]{turney06b}
Peter~D. Turney.
\newblock 2006b.
\newblock Similarity of semantic relations.
\newblock {\em Computational Linguistics}, 32(3):379--416.

\bibitem[\protect\citename{Turney}2008a]{turney08b}
Peter~D. Turney.
\newblock 2008a.
\newblock The latent relation mapping engine: Algorithm and experiments.
\newblock {\em Journal of Artificial Intelligence Research}, 33:615--655.

\bibitem[\protect\citename{Turney}2008b]{turney08a}
Peter~D. Turney.
\newblock 2008b.
\newblock A uniform approach to analogies, synonyms, antonyms, and
  associations.
\newblock In {\em Proceedings of the 22nd International Conference on
  Computational Linguistics (Coling 2008)}, pages 905--912, Manchester, UK.

\bibitem[\protect\citename{Turney}2012]{turney12}
Peter~D. Turney.
\newblock 2012.
\newblock Domain and function: A dual-space model of semantic relations and
  compositions.
\newblock {\em Journal of Artificial Intelligence Research}, 44:533--585.

\bibitem[\protect\citename{Veale}2004]{veale04}
Tony Veale.
\newblock 2004.
\newblock {WordNet} sits the {SAT}: A knowledge-based approach to lexical
  analogy.
\newblock In {\em Proceedings of the 16th European Conference on Artificial
  Intelligence (ECAI 2004)}, pages 606--612, Valencia, Spain.

\bibitem[\protect\citename{Witten \bgroup et al.\egroup }2011]{witten11}
Ian~H. Witten, Eibe Frank, and Mark~A. Hall.
\newblock 2011.
\newblock {\em Data Mining: Practical Machine Learning Tools and Techniques,
  Third Edition}.
\newblock Morgan Kaufmann, San Francisco.

\bibitem[\protect\citename{Zhila \bgroup et al.\egroup }2013]{zhila13}
Alisa Zhila, {Wen-tau} Yih, Christopher Meek, Geoffrey Zweig, and Tomas
  Mikolov.
\newblock 2013.
\newblock Combining heterogeneous models for measuring relational similarity.
\newblock In {\em Proceedings of the 2013 Conference of the North American
  Chapter of the Association for Computational Linguistics: Human Language
  Technologies (NAACL 2013)}, Atlanta, Georgia.

\end{thebibliography}

\end{document}